# PatternGPT: A Pattern-Driven Framework for Large Language Model Text Generation

**Le Xiao, Xin Shan**

College of Information Science and Engineering, Henan University of Technology, Zhengzhou, China

xiaole@haut.edu.cn   shanxin@stu.haut.edu.cn

**Abstract**

Large language models (LLMS) have shown excellent text generation capabilities, capable of generating fluent human-like responses for many downstream tasks. However, applying large language models to real-world critical tasks remains challenging due to their susceptibility to hallucinations and inability to directly use external knowledge. To cope with the above challenges, this paper proposes PatternGPT, a pattern-driven text generation framework for Large Language Models. Firstly, the framework utilizes the extraction capability of Large Language Models to generate rich and diversified structured and formalized patterns, which facilitates the introduction of external knowledge to do the computation, and then draws on the idea of federated learning to use multiple agents to achieve the sharing in order to obtain more diversified patterns, and finally uses judgment criteria and optimization algorithm to search for high-quality patterns to guide the generation of models. Finally, external knowledge such as judgment criteria and optimization algorithms are used to search for high-quality patterns, and the searched patterns are used to guide model generation. This framework has the advantages of generating diversified patterns, protecting data privacy, combining external knowledge, and improving the quality of generation, which provides an effective method to optimize the text generation capability of large language models, and make it better applied to the field of intelligent dialogue and content generation.

**Keywords**: framework; pattern; LLM; text generation

## 1 Introduction

Recently, the field of AI research has undergone a radical change with the advancement of Large Language Models [1-5], especially with the introduction of ChatGPT and GPT-4. These models have demonstrated superior capabilities in dialog, reasoning, and generation by leveraging methods such as thought-chain prompting [6-9] and reinforcement learning based on human feedback (RLHF)[10][11].

However, LLM sometimes suffers from the problem of hallucination [12] during text generation, i.e., the generated text may not correspond to the actual situation [13], resulting in the generated utterances contradicting the facts. There are two main reasons for the hallucination, one is the reason of the training data: since the quality and diversity of the training data are crucial for the performance of LLM, if the training data are biased, erroneous, not complete, or unbalanced, the model may learn wrong semantic relations or generate inaccurate results [14][15], thus generating hallucinations. The second is the reason for training: pre-training a model on a large corpus causes the model to memorize knowledge in its parameters [16][17], and this memorization can help the model to perform better in downstream tasks, but it can also lead to model hallucinations.

In order to reduce the occurrence of hallucinations, some research [18][19][20]have guided LLMs by introducing additional knowledge as prompts to standardize the generative output of LLMs, and the results showed that LLMs incorporating external knowledge achieved favorable generative results. These have structured and formalized knowledge which can be referred to as patterns[21].

When combined with the internal knowledge of the LLM, they enable the LLM to better understand and fulfill the needs of a specific task during the generation process, reduce the occurrence of hallucinations, and produce more reliable and high-quality textual output.

Realizing the significance of patterns for LLM generation, we propose the PatternGPT framework. The framework is pattern-centered. First, diverse and rich patterns are generated through the extraction capability of LLM. Second, the formal characteristics of patterns are utilized and federated learning ideas are borrowed for pattern sharing and exchange to optimize the quality and increase the diversity of patterns. Then, the patterns are filtered using external knowledge such as judgmental criteria to search for appropriate patterns. Finally, the patterns are utilized for model fine-tuning to improve the performance and adaptability of the models, providing a basis for building intelligent dialog systems and content generation systems.

Our contribution is to provide an end-to-end framework for text generation using LLM. It is divided into the following four parts:

**(1) Pattern extraction:** By utilizing the internal knowledge and training experience of LLM, diverse patterns are generated by combining problem understanding, information extraction and linguistic patterns, and the formal and structured nature of the patterns makes it easier to introduce external knowledge for computation.

**(2) Pattern Sharing:** Pattern sharing is performed by utilizing the characteristics of pattern formalization and drawing on the idea of federated learning, which increases the diversity of generated results by sharing patterns generated by different agents, and also shields the data details in the process of sharing to achieve the purpose of partial privacy protection.

**(3) Pattern search and optimization**: introduce external knowledge such as judgment criteria and optimization algorithms to search for high-quality patterns.

**(4) Model fine-tuning**: use the obtained patterns to guide LLM fine-tuning to generate better content.

## 2. Related Work and Background

LLM can generate the answers required by the user based on the knowledge stored in the preprocessing parameters through zero-sample learning, without the requirement of additional parameter updates as well as labeled datasets. However, although LLM shows excellent ability in generating text, its output is less stable [22], sometimes hallucinating, and the output is largely influenced by instructions or prompts[23]. Therefore, in order to improve the performance and stability of LLMs, some research [24] has proposed to retrieve the relevant knowledge of the input problem from the training dataset and use this knowledge as prompt to guide the model generation process. This knowledge retrieval-based approach allows LLMs to better understand and fulfill the needs of a specific task.

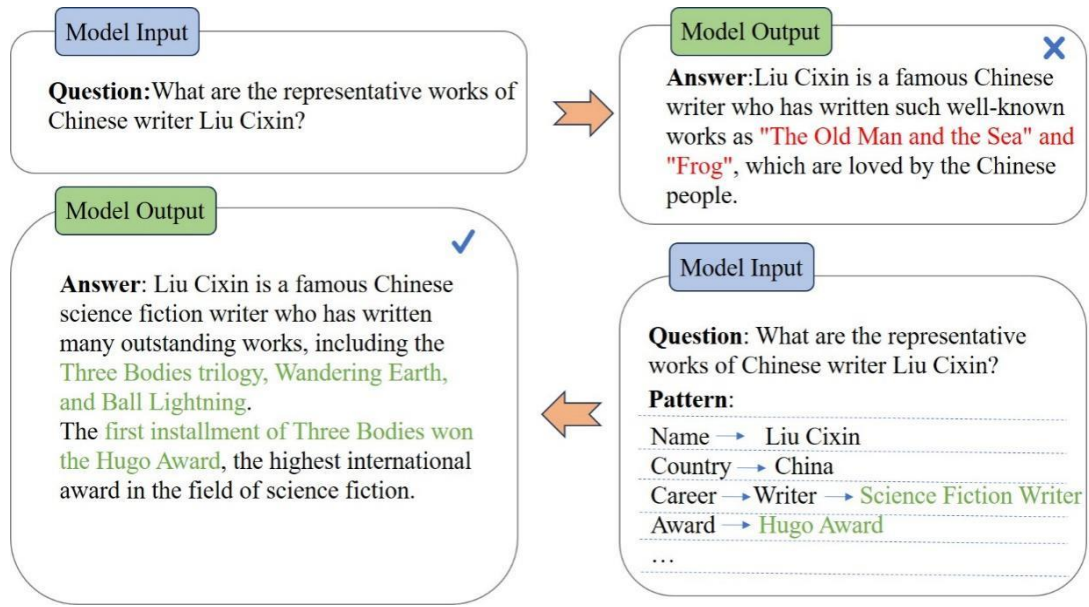

Fig. 1 Patterntic diagram of the effect of patterns as prompts.

Among them, patterns as a kind of knowledge play an important role in providing instructions and prompts. As shown in Figure 1, when we asked "What are the representative works of Chinese writer Liu Cixin?", LLM replied to Hemingway's Old Man and the Sea and Chinese author Mo Yan's Frogs, which is obviously contrary to the fact. When we entered some patterns related to the question into LLM as prompts along with the question, LLM generated reasonable and correct answers, and the patterns "science fiction writer" and "Hugo Award" helped induce LLM to generate works in the science fiction genre. The "science fiction author" and "Hugo Award" in the pattern all help induce LLM to generate science fiction works. This shows the importance of the pattern in guiding LLM to generate output.

### 2.1 Pattern

A pattern refers to a structured, repeatable form of expression or behavior that has a certain rule or template[25]. It can be a textual form, a graphical form, a logical form, or other forms used to represent specific knowledge, information, or ways of acting.

The structured nature of the pattern makes it parsable and computable, which makes it easier to combine external knowledge with the pattern and enables better fine-tuning of LLMs. At the same time, the repeatability and regularity of patterns allow them to be applied to different situations, data, or problems to achieve similar results. By using patterns as instructions or prompts, the style, content, and form of the generated results can be controlled to make them more consistent with the task requirements.

In addition, the abstract nature of patterns makes it possible to extract key features and relationships and ignore unimportant or minor details. This abstraction makes patterns somewhat universal and adaptable, capable of being shared, reused, and optimized. Therefore, the application of patterns can not only promote the performance improvement of LLM but also promote the accumulation and progress of domain knowledge.

### 2.2 Pattern Extraction and Filtering

Pattern extraction aims to automatically identify specific patterns of information from text. It is the basis for tasks such as information extraction, text categorization, and Q&A, and is able to extract useful structured information from large amounts of text data[26].

Previous pattern extraction methods mainly include rule-based methods[27], statistical-based methods[28], hybrid rule- and statistical-based methods[29], sequence-labeling-based methods [30], and deep learning-based methods[31]. Recently, LLMs have demonstrated extraordinary information extraction capabilities[32], and well-structured patterns can be extracted using LLMs [33].LLMs are able to deeply understand and analyze the input text, capturing the semantic, logical, and associative relationships[34]. Compared with traditional extraction methods, LLM has stronger generalization and contextual understanding capabilities and is able to handle complex linguistic expressions and implicit semantic information[35]. On this basis, it is able to extract diverse information and patterns from text [21][36], and the diversity can help us capture information patterns at different levels and perspectives. Different types of patterns may cover different linguistic expressions, logical structures, and semantic relationships, and are able to present issues and knowledge from multiple dimensions. By extracting diverse patterns, we can obtain richer information resources and provide more choices and possibilities for subsequent pattern filtering.

It is necessary to screen the patterns after they are extracted. Among them, diverse patterns play an important role in the pattern filtering process, which can provide more choices and possibilities and enhance the reliability of the patterns. By introducing diverse patterns, we can compare and validate among patterns to find out consistent and reliable patterns [37]. Diverse patterns can complement and corroborate each other to increase the accuracy and credibility of patterns and ensure that the selected patterns are of high quality and relevance to meet specific needs. Pattern filtering can be achieved through manual review, and automated algorithms[38][39][40][41]. In manual review, domain experts or annotators can evaluate and screen patterns based on domain knowledge and experience. Automated algorithms can automatically assess the quality and relevance of patterns based on defined criteria and metrics.

By introducing external knowledge such as reasonable pattern filtering methods and criteria, we can search for the most suitable patterns. By using patterns as instructions and prompts, we can provide more targeted and personalized guidance information during the fine-tuning process of LLM to improve the performance and stability of the model. Through PatternGPT, we can use LLM to extract diverse patterns, making it possible to provide clear guidance for the model, thus promoting the development of text generation, knowledge discovery, and other fields.

**3. PatternGPT**

The framework proposed in this paper is divided into four parts: 1) Pattern extraction, LLM uses internal knowledge and training experience to generate multiple possible patterns based on problem understanding, information extraction, and linguistic patterns. 2) Pattern sharing, exploits the feature of pattern formalizability and draws on the idea of federated learning so that multiple agents can collaborate with each other in the pattern generation task. By sharing patterns generated by different agents in order to increase the diversity of generated results. 3) Pattern selection and generation optimization, improve the performance and performance of LLM in text generation tasks by introducing external knowledge such as judgment criteria and optimization algorithms to search for or generate high-quality patterns. 4) Model fine-tuning, utilize the selected patterns as prompts or generate contextually relevant instructions for model fine-tuning in order to provide more targeted and personalized instruction information to help the language model adapt to a specific task or domain.

Formally, given the set of questions $Q = \{q_1, q_2, ..., q_N\}$, the set of answers $A = \{a_1, a_2, ..., a_N\}$ denotes the standard responses to the corresponding questions and $N$ denotes the total number of questions. Therefore, we need to learn the function $f: Q \rightarrow A$ to map the question descriptions to the corresponding answers. In PatternGPT, we learn the function $f$ by the following steps: we first generate multiple patterns $P_1 = \{p_1, p_2, ..., p_n\}$ from question $q_i$ using LLM, and later combining the patterns generated by other agents $P_2 = \{p_{n+1}, p_{n+2}, ..., p_m\}$ to form the pattern pool $P_0 = \{p_1, p_2, ..., p_n, ..., p_m\}$, then apply the optimization algorithm from the pattern pool $P_0$ to obtain the high-quality pattern $p_{mvp}$, and finally, use $p_{mvp}$ to guide the LLM to generate the answer $a_i$. The structure of the PatternGPT framework is shown in Figure 2.

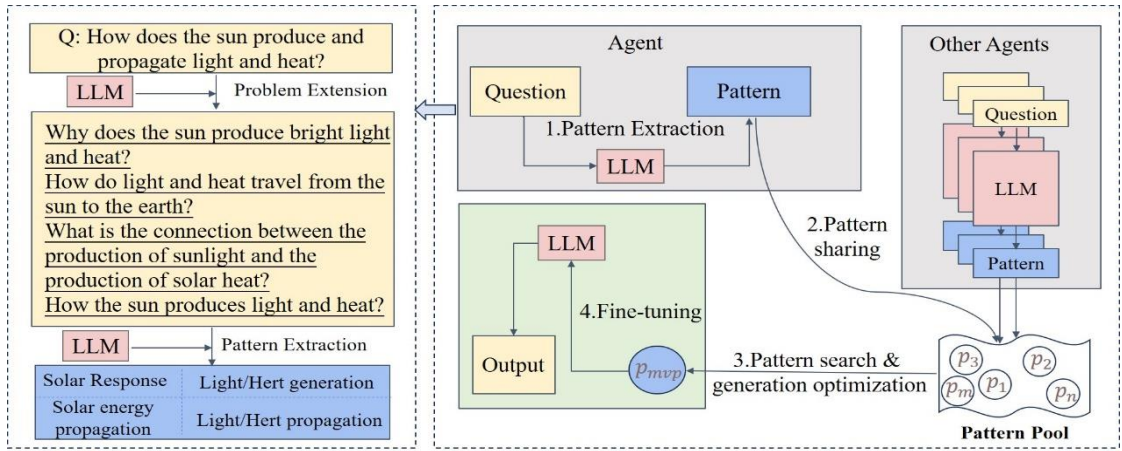

Fig. 2 PatternGPT structure

## 3.1 Pattern Extraction

The core of PatternGPT is to generate a variety of patterns that may contribute to the solution of the problem, and the powerful extraction capabilities of LLM can be utilized to obtain multiple solution patterns for the problem.

### 3.1.1 Problem Input

A user-supplied problem is provided as input to the LLM. e.g. "How does the sun produce and propagate light and heat".

### 3.1.2 Problem Extension

In the problem extension phase, the LLM utilizes its training data and internal semantic understanding capabilities to further extend and expand the knowledge related to the given problem. This extension helps LLM to generate more diverse and rich patterns, even if these patterns are not directly contained in the original problem.

In the process of problem extension, LLM can use the following methods to get diverse problems:

**(1) Variants of similar problems**: LLM can generate variants of problems that extend the coverage of related patterns by changing the formulation, structure, or specific details of the problem.

**(2) Analogies and reasoning**: by applying the ability to draw analogies and reasoning, LLMs can derive new questions related to a problem from known knowledge and semantic relationships.

### 3.1.3 Pattern Generation

In the pattern generation phase, the LLM utilizes its internal knowledge and training experience to generate multiple patterns related to a given problem. This process can be done based on the linguistic patterns and semantic correlations that LLMs have learned in massive amounts of text[33][34].

In the process of generating patterns, LLM can utilize a variety of techniques and strategies in order to increase the diversity of generated patterns. First, LLM can utilize grammatical rules and syntactic structures[42] to generate patterns that conform to grammatical specifications. By following grammatical rules, LLM can generate patterns with different syntactic structures and expressions, thus demonstrating the diversity of problems.

Second, LLM can utilize contextual associations[43] to generate patterns that are semantically relevant to the problem. By deeply understanding the contextual information of the problem, LLM can utilize the learned semantic relevance to generate problem-related patterns. This can make the generated patterns more semantically coherent and able to present the semantic information of the problem from different perspectives.

In addition, LLM can utilize the technique of topic expansion to generate diverse patterns. By analyzing the keywords, topics, contexts, and other factors in the problem, LLM can extend the topic domain of the problem and generate patterns that are related to the problem but cover different aspects. This increases the diversity of the generated patterns and makes them more comprehensive and multi-perspective.

With a deeper understanding of the problem, the LLM can extract relevant information from its memory and knowledge base and generate multiple possible patterns and represent them formally; the formal representation can make the patterns easier to process, compare and match, as well as provide a basis for generation sharing.

### 3.2 Pattern Sharing

In the process of generating patterns, in order to provide a wider range of choices and perspectives, LLM needs to focus on pattern sharing. This means that patterns of different styles, perspectives, or expressions can be shared in a way that satisfies different user needs and preferences. Through the formal representation of patterns, multiple agents can share and exchange generated patterns. Du et al [36] have shown that using multiple agents helps the model to generate more accurate answers. In the process of pattern sharing, our framework can use multiple different LLMs as agents, each of which may have different characteristics, training data, and pre-training parameters, and thus they may have differentiated characteristics and diverse representations in pattern generation. Diverse patterns from different models can be obtained through the sharing of multiple agents, which can be achieved by sharing the data structures of formal pattern representations, using a shared pattern library, or exchanging patterns through a network. Pattern sharing allows multiple agents to benefit from each other by using each other's generation results for further optimization and extension.

In order to realize pattern sharing among multiple agents and to protect the privacy of data details of each agent, we draw on the idea of federated learning to realize pattern generation sharing, specifically:

#### 3.2.1 Establishing a Sharing Mechanism

The pattern-sharing mechanism is the basis for realizing pattern generation sharing among multiple agents, which requires defining communication protocols, privacy protection methods, and so on. The communication protocol can specify the communication methods and interaction rules between agents. The privacy protection mechanism can adopt methods such as differential privacy and secure multi-party computation to ensure that the data privacy between agents is protected.

#### 3.2.2 Agent Selection and Task Assignment

In the multi-agent pattern-sharing process, it is necessary to select the agents that will participate in the pattern generation task and assign the problem to these agents. The selection of agents can be based on several factors, such as performance metrics of the agents, available resources (computing power, storage capacity, etc.), domain expertise, etc. The assignment of tasks needs to take into account the nature and requirements of the problem to ensure that the right agents are involved in the appropriate pattern-generation tasks.

### 3.2.3 Local Pattern Generation

Each agent performs pattern generation in the local environment based on its own local model and data. The agent performs pattern generation based on the problem and generates a collection of patterns related to the problem. These patterns can be obtained based on the generation results of the LLM, the agent's own knowledge and experience, and the communication and collaboration between the agents.

After the local pattern generation is finished, the patterns can be shared among the agents. Pattern sharing can be realized by sending the patterns to a sharing platform or a central server, or through direct communication in a peer-to-peer manner. During the sharing process, data privacy and security can be ensured. Sharing patterns generated by different agents forms a global pattern collection, which improves the quality and diversity of pattern generation and provides a basis for further pattern optimization and search. Overall, pattern generation sharing can bring the following benefits:

**(1) Increase diversity**: by sharing patterns generated by different agents, as much knowledge as possible is acquired to increase the diversity and coverage of the generated results, thus providing richer and more diverse solutions.

**(2) Privacy protection**: data details can be masked during pattern sharing to achieve partial privacy protection.

Through formal representation and generation sharing of patterns, multiple LLMs can collaborate and complement each other in pattern generation tasks, thus providing more powerful and diverse generation capabilities. This combination and sharing approach helps overcome the limitations of a single LLM and provides more comprehensive and flexible options for problem-solving[36].

### 3.3 Pattern Search and Generation Optimization

In the part of pattern search and generation optimization, since LLM itself cannot judge the quality of patterns, to compensate for this defect, we introduce external knowledge such as judgment criteria and optimization algorithms to search or generate high-quality patterns, which determines the quality and applicability of the final generated text. By getting suitable patterns, we can effectively guide LLM for text generation and get more accurate results.

### 3.3.1 Definition of Judgment Criteria

In the optimization phase of pattern search and generation, we objectively and accurately assess the quality of generated patterns through the following judgment criteria.

**(1) Relevance**: measure the relevance of the pattern to the user's problem through semantic similarity or thematic consistency.

**(2) Diversity**: Assessing the differences between patterns in order to avoid generating patterns that are repetitive or excessively similar.

**(3) Syntactic correctness**: Use syntactic analysis techniques to verify whether the syntactic structure of patterns conforms to specifications.

**(4) Contextual coherence**: Ensure that the generated patterns are coherent and reasonable in context.

**(5) Information richness**: assess the amount of information and content richness provided by the pattern.

**3.3.2 Optimal Pattern Search**

We can use algorithms to optimally search the obtained pattern set according to the above judgment criteria to obtain high-quality patterns. Commonly used search methods include intelligent computing algorithms and reinforcement learning[39]. Intelligent computing algorithms such as genetic algorithms[40] and ant colony algorithms[44] can optimize the parameters, structure, or generation of patterns to find the optimal solution by simulating biological evolution or population behavior. Reinforcement learning, on the other hand, is a machine learning method that enables an intelligent body to take optimal actions in a given environment by learning from the interaction of the intelligent body with the environment in order to achieve the goal of maximizing the expected benefits.

These optimization algorithms can help us to obtain high-quality patterns from the generated patterns to guide large language models to generate better content.

**3.3.3 Pattern Aggregation**

In the pattern aggregation phase, we generate more comprehensive, accurate, and rich results by combining multiple patterns. The goal of pattern aggregation is to merge the advantages of different patterns to obtain better generation results, which can be achieved by the following methods:

**(1) Weighted averaging**: for a given set of patterns, a weight can be assigned to each pattern, and then each pattern is multiplied by its corresponding weight and added together to obtain the aggregated patterns. This method is suitable for patterns with similarities or scores, where the weights can be assigned based on the similarities or scores.

**(2) Voting mechanism**: for a given set of patterns, a voting mechanism can be used to determine the aggregated patterns. Each pattern can vote for candidate answers or specific aspects of the generated results, and then the option with the most votes is selected as the aggregated pattern. This approach is suitable for cases where there is some variability between patterns.

**(3) Logic rules**: During pattern aggregation, a set of logic rules can be defined to decide how to combine different patterns. These rules can be based on relationships between patterns, constraints, or other factors to determine the aggregated patterns. The logical rules can be hard rules (e.g., patterns must satisfy specific conditions to be aggregated) or soft rules (e.g., aggregation based on pattern similarity or score).

**(4) Graph or network structure[21]**: patterns can be represented as a graph or network structure, where nodes represent patterns and edges represent relationships or similarities between patterns. Aggregation and integration between patterns can be achieved through the analysis and processing of graph or network structures. For example, similar patterns can be aggregated into the same category using a graph clustering algorithm.

Through pattern aggregation methods, we are able to combine patterns from different sources and with different characteristics to produce more integrated and comprehensive results.

Pattern aggregation improves the accuracy and adaptability of the generated results, and helps to fully utilize the advantages of patterns to enhance the performance and performance of LLM in text generation tasks.

Under the functional framework covering pattern generation and sharing, optimal pattern search, and pattern aggregation, we are able to select the most suitable patterns from a large number of generated patterns. This pattern selection and generation optimization process helps to improve the performance and performance of LLM in text generation tasks. At the same time, these methods also provide the basis and insights for building more intelligent and personalized dialogue systems and content generation systems.

*3.4 Model Fine-tuning*

Selected patterns can be fed back to the LLM as samples or references for fine-tuning, which can provide more targeted and personalized guidance information to help the language model better adapt to a specific task or domain. As shown in Figure 3, by using patterns for fine-tuning, more targeted and personalized guidance information can be provided by generating diverse contextual instructions[45] or prompts[46] to help the language model better adapt to a specific task or domain during the fine-tuning process to a specific task or domain.

Such a fine-tuning approach can improve the performance and generative power of the model, making it more suitable for specific application scenarios and task requirements.

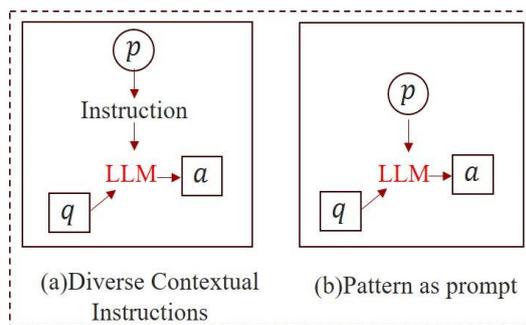

Fig. 3 Model fine-tuning using patterns

### 3.4.1 Diverse Contextual Instructions

By generating diverse contextual instructions, the language model can be guided to focus on specific task requirements or domain knowledge during the fine-tuning process. These instructions can cover different semantic aspects, conditional constraints, and operational requirements to motivate the model to produce more accurate and coherent output. For example, in a text generation task, a series of contextual instructions, including input samples, desired outputs, etc., can be generated to guide the model in fine-tuning the task according to the characteristics of the task type and the target output.

### 3.4.2 Pattern as Prompt

Using patterns as prompts is a common fine-tuning technique. By defining a specific pattern as a prompt, the model can be guided to generate output associated with that pattern. These patterns can be specific questions or instructions, or they can be a set of templated sentence structures. By designing the patterns wisely, the style, content and form of the generated results can be controlled to make them more consistent with the task requirements. At the same time, multiple patterns can be used as prompts to provide more diverse instructional information and increase the model's ability to understand and express the task.

## 4. Summary and Prospect

In this paper, we have introduced a pattern extraction and selection framework, PatternGPT, which aims to address some challenges and problems of LLM in text generation tasks.

In this paper, we introduce PatternGPT, a pattern-driven text generation framework for large language models, aiming to alleviate the hallucination problem of LLM in text generation tasks. By leveraging the internal knowledge and training experience of LLM and combining problem understanding, information extraction, and language pattern generation, PatternGPT is able to generate diverse and useful patterns, which are further increased by the formal representation of patterns and by drawing on the idea of federated learning, where multiple agents can collaborate and complement each other to achieve pattern sharing. Formalized and structured patterns facilitate the introduction of external knowledge for computation and model fine-tuning. By introducing external knowledge such as judgment criteria and optimization algorithms, PatternGPT is able to search for and generate high-quality patterns, which can personalize the guidance of the model generation process, and improve the performance and performance of LLMs in text generation tasks.

However, we are also aware that the current framework has some limitations. First, the pattern extraction stage may be limited by the LLM itself, resulting in less accurate or comprehensive generated patterns. Second, the algorithms and guidelines for pattern selection and generation optimization still have space for improvement and require more in-depth research and practice to improve their effectiveness and efficiency.

In our future research, we will continue to improve our framework and explore more effective pattern extraction methods and optimization algorithms. We will also try to apply the framework to a wider range of tasks and domains to verify its applicability and generalizability. In addition, we plan to further investigate mechanisms for pattern sharing and exchange to build a more robust and sustainable pattern repository.

Overall, our research provides an effective PatternGPT for the application of LLM in text-generation tasks. through diverse pattern generation, optimized pattern selection, and personalized guidance information, we believe that the framework will provide a basis for building intelligent dialogue systems and content generation systems, driving further development and applications in the field of AI research.